%% file: root.tex
\documentclass[letterpaper, 10 pt, conference]{ieeeconf}  

\IEEEoverridecommandlockouts                              

\usepackage{graphicx} 
\usepackage{tabularx}
\usepackage{stfloats}

\usepackage{cite}
\usepackage{algorithm}
\usepackage{algorithmic}
\usepackage{amsfonts}
\usepackage{amsmath}
\usepackage{array,booktabs,arydshln}
\usepackage{bbm}
\usepackage{subfig}
\usepackage{booktabs}
\usepackage{float}
\usepackage[inkscapelatex=false]{svg}
\usepackage{xcolor}
\usepackage{array}
\usepackage{multirow}
\usepackage{flushend}

\definecolor{green}{RGB}{11,155,13}

\title{\LARGE \bf
Reward Training Wheels: Adaptive Auxiliary Rewards for \\Robotics Reinforcement Learning}

\author{Linji Wang$^{*}$, Tong Xu$^{*}$, Yuanjie Lu, and Xuesu Xiao 
\thanks{All authors are with the Department of Computer Science, George Mason University {\tt\scriptsize \{lwang44, txu25, ylu22, xiao\}@gmu.edu}}
\thanks{*Equally contributing authors}
}

\begin{document}

\maketitle
\thispagestyle{empty}
\pagestyle{empty}

\begin{abstract}
Robotics Reinforcement Learning (RL) often relies on carefully engineered auxiliary rewards to supplement sparse primary learning objectives to compensate for the lack of large-scale, real-world, trial-and-error data. While these auxiliary rewards accelerate learning, they require significant engineering effort, may introduce human biases, and cannot adapt to the robot's evolving capabilities during training. In this paper, we introduce Reward Training Wheels (RTW), a teacher-student framework that automates auxiliary reward adaptation for robotics RL. 
To be specific, the RTW teacher dynamically adjusts auxiliary reward weights based on the student's evolving capabilities
to determine which auxiliary reward aspects require more or less emphasis to improve the primary objective. 
We demonstrate RTW on two challenging robot tasks: navigation in highly constrained spaces and off-road vehicle mobility on vertically challenging terrain. In simulation, RTW outperforms expert-designed rewards by 2.35\% in navigation success rate and improves off-road mobility performance by 122.62\%, while achieving 35\% and 3X faster training efficiency, respectively. 
Physical robot experiments further validate RTW's effectiveness, achieving a perfect success rate (5/5 trials vs. 2/5 for expert-designed rewards) and improving vehicle stability with up to 47.4\% reduction in orientation angles.

\end{abstract}

\input{contents/intro.tex}
\input{contents/related.tex}

\input{contents/approach.tex}

\input{contents/experiments.tex}
\input{contents/conclusions.tex}

\bibliographystyle{IEEEtran}
\bibliography{IEEEabrv,references}

\end{document}

%% file: contents/intro.tex
\section{Introduction}
\label{sec:intro}
Designing robust and sample-efficient Reinforcement Learning (RL) methods for real-world robotics remains an enduring challenge. Whether the goal is to enable a ground robot to navigate narrow corridors or to drive an off-road vehicle up steep, uneven terrain, reward functions are pivotal to guide an agent's behavior. In principle, one could rely on sparse, goal-only rewards—for example, awarding a single bonus upon successfully reaching a target. However, such sparse feedback typically proves insufficient for complex robotic tasks, especially considering real-world data scarcity, as the agent has little incentive to avoid obstacles or maintain safe maneuvers, making it difficult to reach the goal and receive the sparse reward through trial and error~\cite{ng1999policy, hadfield2017inverse}.

To address this limitation, roboticists introduce auxiliary rewards—terms that penalize collisions, encourage partial progress, or maintain safety constraints. While these auxiliary signals can dramatically accelerate learning, they also multiply the engineering burden: one must decide which auxiliary terms to include and how much to weigh each one, often through human trial and error and still leading to biased reward design~\cite{dewey2014reinforcement, lu2023leveraging, devidze2024informativeness}. Worse yet, an auxiliary reward that benefits early exploration (e.g., collision penalties to teach cautious movement) can become counterproductive later, as the robot may overly prioritize always staying unnecessarily far from obstacles rather than completing the task efficiently.

Recent works in robotic RL reveal the extent of this auxiliary reward design burden. Tasks like navigation in highly constrained spaces~\cite{xiao2022autonomous, xiao2023autonomous, xiao2024autonomous} commonly include several hand-tuned rewards (e.g., distance-to-goal bonuses, collision penalties, velocity constraints, or time penalties), carefully balanced by human intuition~\cite{perille2020benchmarking, xu2023benchmarking, xu2021machine}. Similarly, off-road mobility on vertically challenging terrain often relies on auxiliary terms to mitigate unsafe maneuvers—penalizing rollovers or slipping—while rewarding forward progress~\cite{xu2024reinforcement}. In each case, manually engineering these auxiliary signals can be time-consuming and prone to human biases and overfitting a particular training phase. These challenges highlight the need for a principled approach that can adapt auxiliary rewards throughout the training process.

Imagine training wheels used to learn how to ride a bicycle.  An effective human teacher uses training wheels at the beginning to guide learning in a safe manner, encourages the student to depend less and less on them as learning progresses, and eventually completely remove them. 
Inspired by the diminishing role played by the training wheels, we hypothesize that RL auxiliary rewards also need to adapt to the student capabilities.  
In this paper, we introduce Reward Training Wheels (RTW), a teacher-student framework that preserves the robot's ultimate task objective (learning to ride a bicycle) while dynamically adapting auxiliary components (reducing the reliance on the training wheels). The teacher agent monitors key training signals—such as success rate and other performance metrics—to assess the student robot's current capability and adjust auxiliary reward weights accordingly. As a result, the student leverages RTW to improve learning efficiency and performance through the adaptive auxiliary rewards throughout the learning process.

We apply RTW on two difficult robotic tasks: (1) navigation in highly constrained  spaces and (2) off-road vehicle mobility on vertically challenging terrain. Compared to expert-designed auxiliary rewards, our approach outperforms by 2.35\% in navigation success rate and improves off-road mobility performance by 122.62\%, while achieving 35\% and 3X faster training efficiency, respectively. 
Physical robot experiments further validate RTW's effectiveness, achieving a perfect success rate (5/5 trials vs. 2/5 for expert-designed rewards) and improving vehicle stability with up to 47.4\% reduction in orientation angles.

%% file: contents/related.tex
\section{Related Work}
\label{sec::related}
RTW is related with curriculum learning and reward shaping, for which we review related work in this section.

\subsection{Curriculum Learning}
Curriculum learning, first formalized by Bengio et al.~\cite{bengio2009curriculum}, has become increasingly important in RL as a strategy to improve sample efficiency and final performance. Traditional curriculum approaches in robotics focus on progressively increasing task difficulty~\cite{narvekar2020curriculum, florensa2018automatic, wang2024grounded}, often by modifying environmental parameters such as obstacle density~\cite{ivanovic2019barc} or terrain complexity~\cite{hwangbo2019learning}. Recent work by Freitag et al.~\cite{freitag2024sample} demonstrated that well-designed curricula can significantly enhance sample efficiency in robotic navigation tasks. Similarly, Mehta et al.~\cite{mehta2020active} employed a curriculum-based approach for learning complex manipulation skills by progressively introducing more challenging scenarios.  While effective, these approaches introduce additional engineering complexity by requiring multiple task variants and can lead to distribution shifts when transferring to the target environment. Unlike these methods that primarily modify task configurations or initial states, our teacher-student framework maintains a consistent environment while dynamically adapting the auxiliary reward structure as the student progresses.

\subsection{Reward Shaping}
Reward shaping has long been recognized as a critical element in RL~\cite{ng1999policy, hadfield2017inverse}. Potential-based reward shaping~\cite{ng1999policy} provides theoretical guarantees for preserving optimal policies, though how to design potential functions to efficiently lead to optimality remains challenging. Devlin and Kudenko~\cite{devlin2012dynamic} extended this to dynamic potential-based reward shaping, allowing the shaping function to change over time but still requiring domain knowledge for potential function design.
Methods for automatic reward design have emerged as alternatives to manual engineering. Singh et al.~\cite{singh2010intrinsically} explored intrinsically motivated RL, where auxiliary rewards emerge from the agent's learning progress. More recently, Devidze et al.~\cite{devidze2024informativeness} proposed metrics to evaluate the informativeness of reward signals during training. These approaches recognize the limitations of fixed auxiliary rewards but typically do not adapt rewards based on the agent's current capabilities as RTW does.

Several recent studies have explored adaptive rewards in robotic RL. Gupta et al.~\cite{gupta2022unpacking} investigated how different reward formulations affect robotic manipulation learning, showing that reward design choices significantly impact training dynamics. Riedmiller et al.~\cite{riedmiller2018learning} introduced a learning from play framework that separates reward specification from policy learning, allowing more natural skill acquisition.
In the context of mobile robots, Xu et al.~\cite{xu2023benchmarking} benchmarked different reward structures for navigation tasks, while Xu et al.~\cite{xu2024verti} explored reward design for off-road mobility. These studies highlight the challenge of creating effective reward functions for complex robotic tasks but typically rely on static reward formulations. Our approach builds upon these insights by introducing automatic adaptation of auxiliary rewards throughout the training process.

%% file: contents/approach.tex
\section{Approach}
\label{sec:approach}

\begin{figure}[t]
    \centering
    \includegraphics[width=\columnwidth]{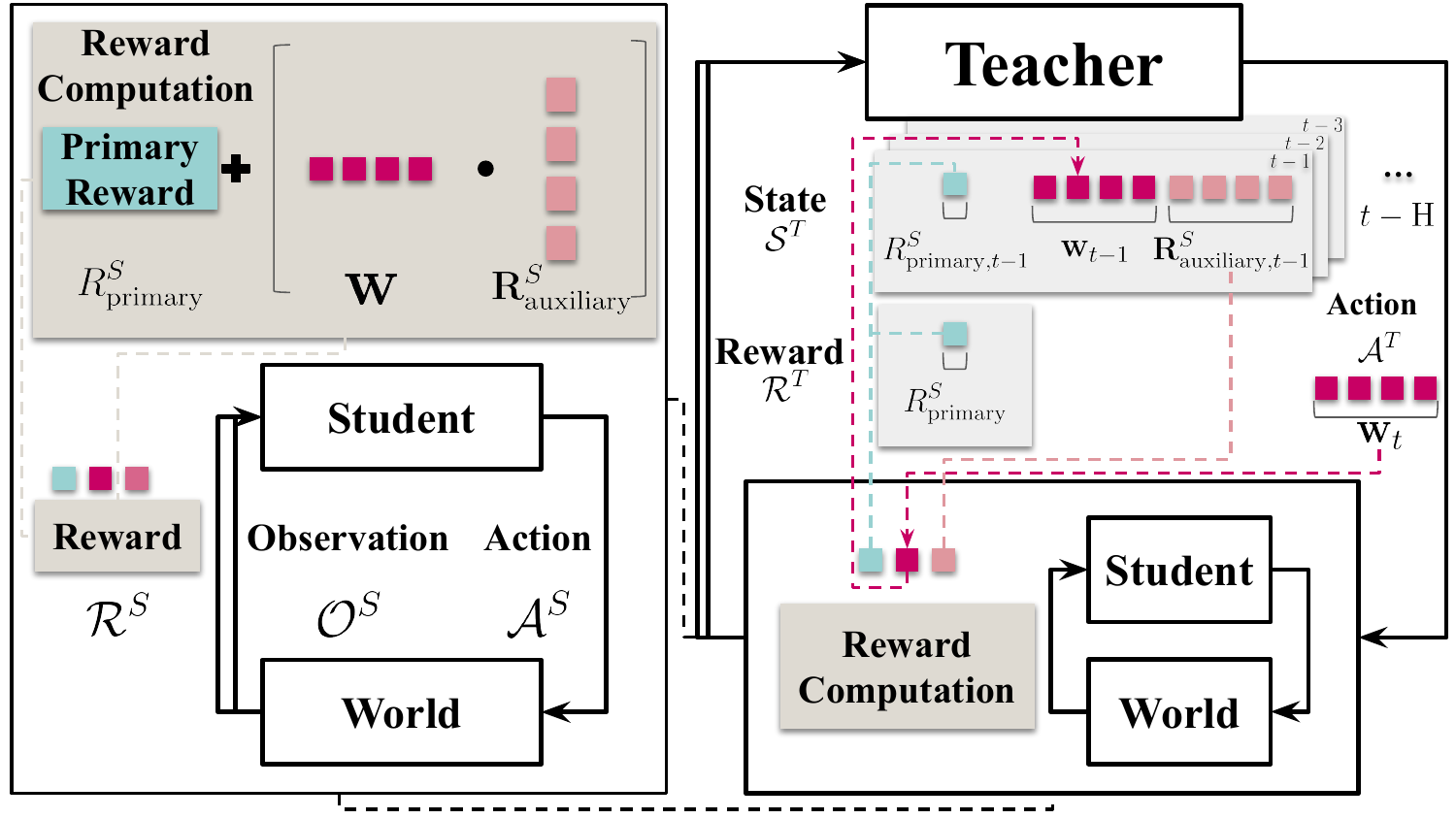}
    \caption{Overview of Reward Training Wheel: (Left) The student agent interacts with the world and receives a reward composed of a primary component and weighted auxiliary components. (Right) The teacher agent maintains the history of previous weights, primary rewards, and auxiliary rewards as its state, and generates new weights as its action to optimize the student's learning process.}
    \label{fig:approach}
    \vspace{-10pt}
\end{figure}

RTW is a dual-agent framework for dynamic auxiliary reward design in robotics RL. As depicted in Fig.~\ref{fig:approach}, our framework employs a hierarchical structure comprising two specialized agents: 
\begin{itemize}
    \item  A Partially Observable Markov Decision Process (POMDP) for the student agent ($\cdot^S$).
    \item A Markov Decision Process (MDP) for the teacher agent ($\cdot^T$) generating auxiliary reward weights.
\end{itemize}

\subsection{Student Agent}
\label{subsec:student_agent}
The student agent operates in a POMDP defined as $\mathcal{M}^S = \langle \mathcal{S}^S, \mathcal{A}^S, \mathcal{O}^S, \mathcal{T}^S, \Omega^S, \mathcal{R}^S, \gamma^S \rangle$, where $\mathcal{S}^S$ is the state space of the robotic task, $\mathcal{A}^S$ is the robot action space, and $\mathcal{O}^S$ is the robot observation space. The transition function $\mathcal{T}^S: \mathcal{S}^S \times \mathcal{A}^S \rightarrow \mathcal{S}^S$ defines state dynamics, while the observation function $\Omega^S: \mathcal{S}^S \times \mathcal{A}^S \rightarrow \mathcal{O}^S$ determines what the agent can perceive. The reward function $\mathcal{R}^S: \mathcal{S}^S \times \mathcal{A}^S \times \mathcal{S}^S \times \mathbf{w} \rightarrow \mathbb{R}$ is parameterized by auxiliary reward weights $\mathbf{w}$, and $\gamma^S \in [0, 1]$ is the discount factor. The student agent's goal is to learn a policy $\pi^S: \mathcal{O}^S \rightarrow \mathcal{A}^S$ that maximizes the expected cumulative reward. The student agent's goal is to learn a policy $\pi^S: \mathcal{O}^S \rightarrow \mathcal{A}^S$ that maximizes the expected cumulative reward.

\subsubsection{Decomposed Reward Structure}
We decompose the student's reward function into a primary reward component and multiple auxiliary components:
\begin{equation}
\mathcal{R}^S(s, a, s', \mathbf{w}) = R^S_{\text{primary}}(s, a, s') + \sum_{k=1}^{K} w_k \cdot \mathbf{R}^S_{\text{aux},k}(s, a, s')
\end{equation}
where:
\begin{itemize}
\item $R^S_{\text{primary}}(s, a, s')$ is the primary reward (e.g., successful task completion),
\item $\mathbf{R}^S_{\text{aux},k}(s, a, s')$ is the $k$-th auxiliary reward component, and
\item $w_k \in [0,1]$ is the weight generated by the teacher agent
\end{itemize}

For each component, the teacher dynamically generates a weight $w_k$ that determines the importance of that auxiliary reward during the learning progress.

\subsection{Teacher Agent}
\label{subsec:teacher_agent}
The teacher agent operates in a fully observable Markov Decision Process (MDP) defined as $\mathcal{M}^T = \langle \mathcal{S}^T, \mathcal{A}^T, \mathcal{T}^T, \mathcal{R}^T, \gamma^T \rangle$, where $\mathcal{S}^T$ is the teacher state space containing the history of student performance and reward information. The teacher's action space $\mathcal{A}^T = [0,1]^K$ represents the normalized auxiliary reward weights $\mathbf{w}$, with $K$ being the number of auxiliary reward components. The transition function $\mathcal{T}^T: \mathcal{S}^T \times \mathcal{A}^T \rightarrow \mathcal{S}^T$ governs state dynamics, while the reward function $\mathcal{R}^T: \mathcal{S}^T \times \mathcal{A}^T \times \mathcal{S}^T \rightarrow \mathbb{R}$ is based on student task performance. The discount factor $\gamma^T \in [0, 1]$ determines the importance of future rewards.

The teacher's state $s_t^T \in \mathcal{S}^T$ at time $t$ comprises three main components with a history horizon $H$:

\begin{equation}
s_t^T = \{\mathbf{w}_{t-H:t-1}, \mathbf{p}_{t-H:t-1}, \mathbf{r}_{t-H:t-1}\}
\end{equation}
where:
\begin{itemize}
    \item $\mathbf{w}_{t-H:t-1} = \{\mathbf{w}_i\}_{i=t-H}^{t-1}$ is the history of auxiliary reward weight configurations, 
\item $\mathbf{p}_{t-H:t-1} = \{p_i\}_{i=t-H}^{t-1}$ is the history of primary rewards, and
    \item $\mathbf{r}_{t-H:t-1} = \{\mathbf{r}_i^k\}_{i=t-H,k=1}^{t-1,K}$ is the history of auxiliary reward vectors.
\end{itemize}

This state representation allows the teacher to track three critical aspects of the student's learning process: (1) what weight configurations have been assigned previously, (2) how well the student performs overall on the primary objective, and (3) how the student performs with respect to each auxiliary reward component. By monitoring these aspects, the teacher can determine which auxiliary rewards to emphasize or de-emphasize as the student learning progresses.

\subsection{Learning Process}
\label{subsec:learning_process}

\subsubsection{Student Learning}
The student agent learns to perform the task using RL (e.g., PPO~\cite{schulman2017proximal}) in the environment to maximize the expected cumulative reward:
\begin{equation}
J^S(\pi^S_{\theta^S}, \mathbf{w}) = \mathbb{E}_{\tau^S \sim \pi^S_{\theta^S}}\left[\sum_{t=0}^{T} {(\gamma^S)}^t \mathcal{R}^S(s_t, a_t, s_{t+1}, \mathbf{w}_t)\right]
\label{eq:student_objective}
\end{equation}
where $\tau^S = (s_0, a_0, s_1, a_1, ..., s_T)$ is a trajectory sampled from the student policy $\pi^S_{\theta^S}$, parameterized by $\theta^S$.
At each training iteration, the student collects trajectories using the current policy and reward weights and then updates its policy parameters:
\begin{equation}
\theta^S \gets \theta^S + \eta^S \nabla_{\theta^S} J^{S}(\pi^S_{\theta^S}, \mathbf{w})
\end{equation}
where $\eta^S$ is the learning rate.
\begin{algorithm}[tbp]
\small
\caption{Reward Training Wheels}
\label{alg:aux_reward_design}
\begin{algorithmic}[1]
\STATE \textbf{Input:} Initial parameters $\theta^S$ and $\theta^T$, learning rates $\eta^S$ and $\eta^T$, primary reward $R^S_{\text{primary}}$, auxiliary reward components $\{\mathbf{R}^S_{\text{aux},k}\}_{k=1}^K$,  history horizon $H$, number of iterations $N$
\STATE \textbf{Output:} Trained student policy $\pi^S_{\theta^S}$
\STATE Initialize student policy $\pi^S_{\theta^S}$, teacher policy $\pi^T_{\theta^T}$, and teacher state $s_0^T$
\FOR{$t = 0$ to $N-1$}
    \STATE Generate auxiliary reward weights $\mathbf{w}_t = \pi^T_{\theta^T}(s_t^T)$
    \STATE Formulate student reward $\mathcal{R}^S = R^S_{\text{primary}} + \sum_{k=1}^{K} w_{t,k} \cdot \mathbf{R}^S_{\text{aux},k}$

    \STATE Collect student trajectories $\{\tau_i^S\}_{i=1}^M$ using current policy and reward function
    \STATE Compute performance metrics $\mathbf{p}_t$ and auxiliary reward values $\mathbf{r}_t$
    \STATE Update student policy: $\theta^S \gets \theta^S + \eta^S \nabla_{\theta^S} J^{S}(\pi^S_{\theta^S}, \mathbf{w})$
    \STATE Update teacher state with history horizon: $s_{t+1}^T = \{\mathbf{w}_{t-H+1:t}, \mathbf{p}_{t-H+1:t}, \mathbf{r}_{t-H+1:t}\}$
    \STATE Update teacher policy: $\theta^T \gets \theta^T + \eta^T \nabla_{\theta^T} J^{T}(\pi^T_{\theta^T}; \mathbf{w})$
\ENDFOR
\RETURN $\pi^S_{\theta^S}$
\end{algorithmic}
\end{algorithm}
\subsubsection{Teacher Learning}
The teacher agent monitors the student's performance metrics to design adaptive auxiliary reward structures. 
The student performance history combined with reward history reveals what specific aspects of the task the student  struggles with and masters. 
Based on this assessment, the teacher generates appropriate auxiliary reward weights $\mathbf{w}_t$ that emphasize aspects that need improvement while reducing emphasis on what has already been mastered.

To be specific, the teacher's objective is to maximize the student's task performance, which we define as:

\begin{equation}
J^T(\pi^T_{\theta^T}; \mathbf{w}) = \mathbb{E}_{\tau^T \sim \pi^T_{\theta^T}}\left[\sum_{t=0}^{T} \gamma^T_t (r^T_t; \mathbf{w}_t)\right]
\end{equation}
where $\tau^T = (s_0^T, a_0^T, s_1^T, a_1^T, ..., s_T^T)$ is a trajectory in the teacher's MDP, and $r^T_t$ is the teacher's reward at time $t$, defined as the student's primary reward under the current auxiliary reward weights $\mathbf{w}_t$.

We implement the teacher using a separate RL agent that observes the student's performance metrics and outputs auxiliary reward weights. The teacher's policy is updated using:

\begin{equation}
\theta^T \gets \theta^T + \eta^T \nabla_{\theta^T} J^T(\pi^T_{\theta^T}; \mathbf{w})
\end{equation}
where $\eta^T$ is the teacher's learning rate.

\subsection{Training Algorithm}
\label{subsec:training_algorithm}
Algorithm~\ref{alg:aux_reward_design} summarizes our training procedure. We initialize both student and teacher policies (line 3). For each training iteration, the teacher generates auxiliary reward weights based on its current state (line 5), which are used to form the student's reward function (line 6). The student collects trajectories using this reward function (line 7), and performance metrics are calculated (line 8). The student policy is then updated (line 9), and the teacher's state is updated to include the latest weight configuration and performance metrics (line 10). Finally, the teacher policy is updated based on the student's performance (line 11). This process continues until convergence.

%% file: contents/experiments.tex
\section{Experiments and Results}
\label{sec:experiments}
\begin{figure}[ht]
    \centering
    \includegraphics[width=\columnwidth]{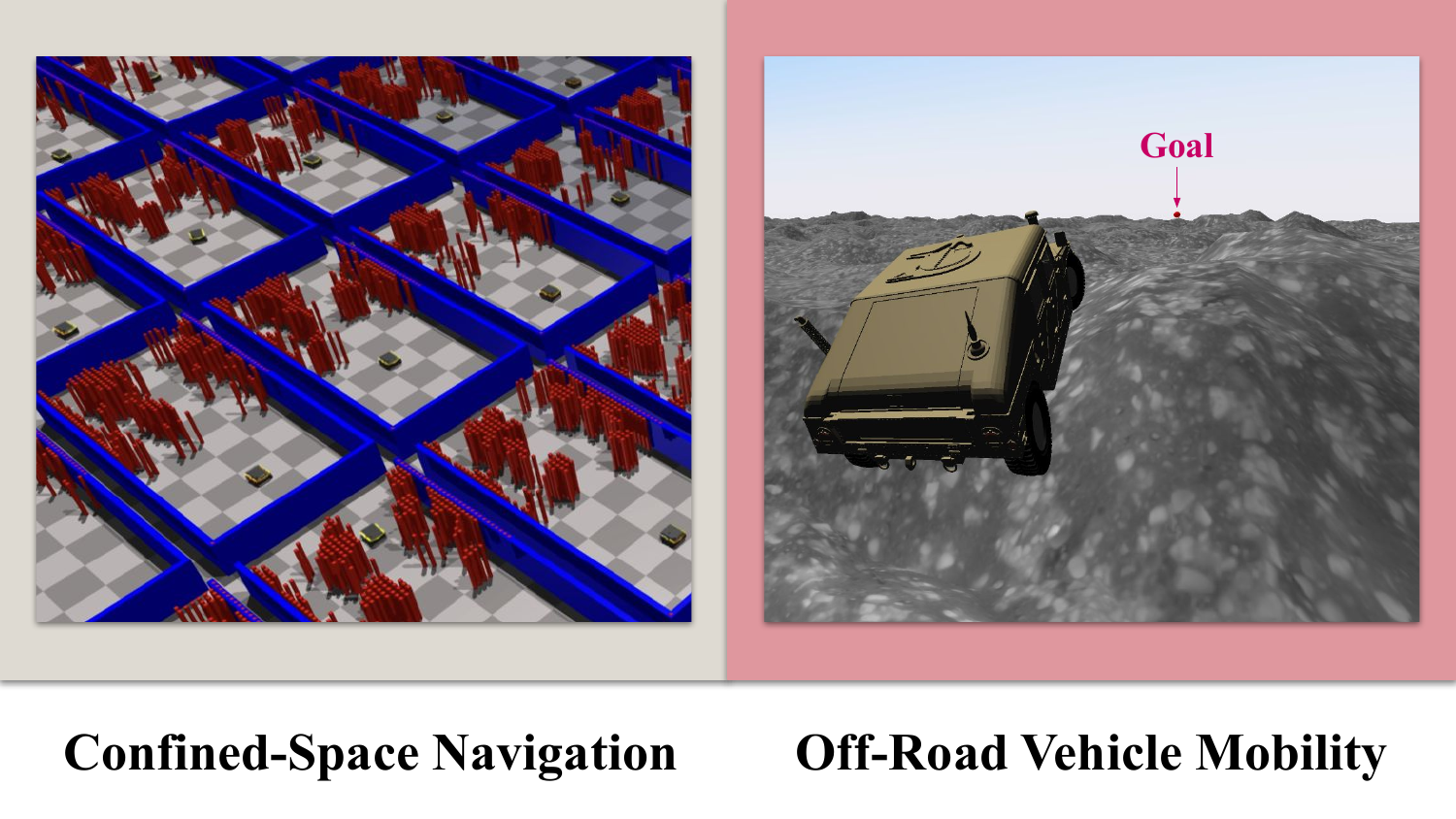}
    \caption{Simulation environments for the two robotics tasks: (Left) Confined-Space Navigation, requiring precise maneuvering through narrow corridors. (Right) Off-Road Vehicle Mobility, challenging the agent to traverse uneven terrain to reach a goal.}
    \label{fig::sim}
    \vspace{-10pt}
\end{figure}

In this section, we evaluate RTW’s effectiveness in dynamic auxiliary reward design by comparing it against three baselines: Expert-designed Reward (ER), which uses fixed auxiliary weights; Reward Randomization (RR), where auxiliary weights change randomly; and Model Predictive Path Integral (MPPI)~\cite{williams2017model}, a conventional planner. We test on two challenging robotic domains: Confined-Space Navigation and Off-Road Vehicle Mobility (Fig.~\ref{fig::sim}). We also conduct physical experiments to validate our approach in real-world settings. Here, we describe our experimental setup, evaluation metrics, and results.
\begin{figure}[tp]
    \centering
    \includegraphics[width=\columnwidth]{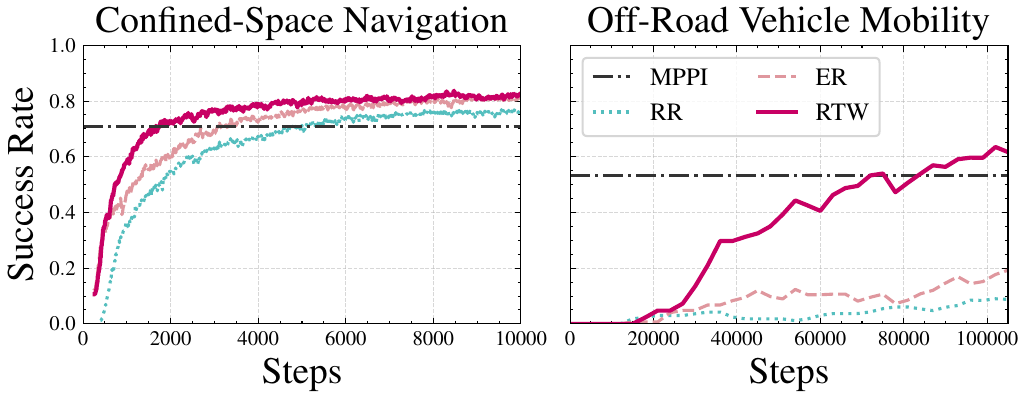}
    \caption{Learning curves showing success rate vs. training steps for MPPI, RR, ER, and RTW. Left: Confined-Space Navigation task. Right: Off-Road Vehicle Mobility task. RTW achieves higher succcess rate with fewer training steps in both tasks.}

    \label{fig:training_efficiency}
    \vspace{-10pt}
\end{figure}

\subsection{Simulation Experiments}
\label{subsec:sim_setup}
\subsubsection{Experimental Setup}
We conduct simulation experiments in two domains: (1) Confined-Space Navigation and (2) Off-Road Vehicle Mobility. 
\paragraph{Confined-Space Navigation.}
The environment consists of randomly generated maps with narrow corridors and obstacles. The robot navigates from a starting position to a target location without collisions. The reward components include:
\begin{itemize}
\item $\mathcal{R}^S_{\text{primary}}$: +20.0 for reaching the target, 
\item $\mathcal{\mathbf{R}}^S_{\text{aux},1}$: -2.0 for collisions with obstacles,
\item $\mathcal{\mathbf{R}}^S_{\text{aux},2}$: +2.0 for progress distance to the goal, and
\item $\mathcal{\mathbf{R}}^S_{\text{aux},3}$: +1.0 for maintain safe distance from obstacles.
\end{itemize}

\paragraph{Off-Road Vehicle Mobility}
The environment features procedurally generated terrain with varying slopes, roughness, and obstacles. The vehicle must reach a target while maintaining stability. The reward components include:
\begin{itemize}
\item $\mathcal{R}^S_{\text{primary}}$: +1000.0 for reaching the target,
\item $\mathcal{\mathbf{R}}^S_{\text{aux},1}$: +5.0 for progress distance to the goal,
\item $\mathcal{\mathbf{R}}^S_{\text{aux},2}$: +1.0 for vehicle current speed (up to maximum speed - 4m/s),
\item $\mathcal{\mathbf{R}}^S_{\text{aux},3}$: -0.5 for vehicle stalling or lack of progress, and
\item $\mathcal{\mathbf{R}}^S_{\text{aux},4}$: -0.5 for exceeding roll and pitch threshold angles.
\end{itemize}

\paragraph{Implementation Details}
We implement RTW and the RL baselines using PPO with the hyperparameters in Table~\ref{tab:hyperparameters}. For MPPI, we use standard parameters with a planning horizon of 2.0 seconds.

\begin{table}[tp]
\centering
\caption{Hyperparameters for RTW Implementation.}
\label{tab:hyperparameters}
\renewcommand{\arraystretch}{1.1}
\begin{tabular}{lll}
\toprule[1pt]
 & \textbf{Navigation} & \textbf{Mobility} \\
\midrule
\textbf{Teacher} & & \\
Learning Rate & 3e-4 & 3e-4 \\
PPO Epochs & 10 & 10 \\
Size of State Space & 7 & 9 \\
Size of Action Space & 3 & 4 \\
Episodes per Update & 10 & 2 \\ 
\midrule
\textbf{Student} & & \\
Learning Rate & 3e-4 & 5e-4 \\
PPO Epochs & 10 & 5 \\
Batch Size & 64 & 1500 \\
Timesteps/iter & 256 & 3000 \\
Policy Network & [512, 512] & [64, 64] \\
Value Network & [512, 512] & [64, 64] \\
Number of Environments & 128 & 5 \\
\bottomrule
\end{tabular}
\vspace{-10pt}
\end{table}

\paragraph{Evaluation Metrics.}
For Confined-Space Navigation, we measure: (i) Success Rate (percentage of trials reaching goal without collisions), (ii) Mean Traversal Time (seconds per successful trial), (iii) Average Distance to Goal (final distance in meters for all trials), and (iv) Average Speed (meters per second, with a maximum of 2 m/s).

For Off-Road Vehicle Mobility, we track: (i) Success Rate out of 30 trials (percentage of reaching goal without rollover or getting stuck), (ii) Mean traversal time (of successful trials in seconds), (iii) Average roll/pitch angles with variance (in degrees).

\subsubsection{Simulation Results}
\label{subsec:sim_results}
\paragraph{Performance Comparison}
Table~\ref{tab:main_results} presents quantitative results across both domains. RTW consistently outperforms all baselines in every metric.

\begin{table*}[tb]
\centering
\caption{Performance Comparison on Confined-Space Navigation and Off-Road Vehicle Mobility (mean $\pm$ std).
$\uparrow$ indicates higher is better; $\downarrow$ indicates lower is better.}
\label{tab:main_results}
\resizebox{\textwidth}{!}{%
\renewcommand{\arraystretch}{1.2}
\begin{tabular}{lcccccccc}
\specialrule{1.2pt}{0pt}{0pt}
& 
\multicolumn{4}{c}{\centering \textbf{Confined-Space Navigation}} & 
\multicolumn{4}{c}{\centering \textbf{Off-Road Vehicle Mobility}} \\
\cmidrule(lr){2-5} \cmidrule(lr){6-9}
\textbf{Method}
& \textbf{Success} $\uparrow$ 
& \textbf{Time} $\downarrow$ 
& \textbf{Dist. to Goal} $\downarrow$ 
& \textbf{Avg. Speed} $\uparrow$ 
& \textbf{Success} $\uparrow$
& \textbf{Time} $\downarrow$ 
& \textbf{Absolute Roll} $\downarrow$ 
& \textbf{Absolute Pitch} $\downarrow$ \\
\midrule
ER        
& 80.32 $\pm$ 1.98 
& 6.23 $\pm$ 0.65 
& 1.52 $\pm$ 0.19 
& 1.24 $\pm$ 0.43 
& 34.44 $\pm$ 9.56
& 31.68 $\pm$ 0.83
& 5.47 $\pm$ 0.08
& 5.33 $\pm$ 0.06
\\
RR      
& 76.15 $\pm$ 2.74 
& 6.89 $\pm$ 0.97 
& 1.76 $\pm$ 0.37 
& 1.12 $\pm$ 0.31
& 12.22 $\pm$ 6.85
& 30.43 $\pm$ 1.25
& 5.16 $\pm$ 0.35
& \textbf{5.12} $\pm$ 0.13
\\
MPPI          
& 71.01 $\pm$ 2.21 
& 6.10 $\pm$ 1.02 
& 1.89 $\pm$ 0.47 
& 0.91 $\pm$ 0.36
& 53.33 $\pm$ 2.03
& 30.26 $\pm$ 1.54
& \textbf{5.10} $\pm$ 16.00
& 6.51 $\pm$ 35.39
\\
\textbf{RTW (Ours)}
& \textbf{82.67} $\pm$ 1.95
& \textbf{6.02} $\pm$ 0.86
& \textbf{1.37} $\pm$ 0.17
& \textbf{1.32} $\pm$ 0.28
& \textbf{76.67} $\pm$ 2.72
& \textbf{29.39} $\pm$ 0.56
& 5.52 $\pm$ 0.17
& 6.06 $\pm$ 0.37
\\
\specialrule{1.2pt}{0pt}{0pt}
\end{tabular}}
\end{table*}

In the Confined-Space Navigation task, RTW achieves the highest success rate among all approaches (82.67\%) and outperforms baselines across all metrics. MPPI shows the lowest success rate, highlighting the difficulty of this navigation task for conventional planning approaches. ER performs reasonably well with expert-designed weights, while RR's inferior performance demonstrates that merely randomizing auxiliary reward weights is detrimental. This contrast between RR and RTW underscores our teacher agent's effectiveness in discovering strategic reward adaptations rather than arbitrary changes.
For Off-Road Vehicle Mobility, RTW demonstrates significantly improvements with a 76.67\% success rate, outperforming the next best MPPI approach. This represents a 43.76\% relative improvement over the conventional planning approach and a 122.62\% improvement over the ER baseline.  RTW also achieves the fastest traversal time, indicating its ability to navigate efficiently through vertically challenging terrain. While MPPI and RR show slightly better absolute roll and pitch metrics, respectively, these come at the cost of substantially lower success rates, suggesting overly conservative navigation strategies that fail to reach goals. RTW maintains reasonable vehicle stability metrics while successfully completing far more trajectories, demonstrating its ability to balance multiple competing objectives in complex off-road environments.

\paragraph{Training Efficiency}
Fig.~\ref{fig:training_efficiency} presents the learning curves for all methods. RTW consistently achieves high performance with fewer training steps than the RL baselines. In the navigation task, RTW reaches an 80\% success rate approximately 35\% faster than ER, while RR fails to reach this threshold. In the off-road mobility task, RTW demonstrates remarkable sample efficiency, reaching the same performance threshold approximately 3X faster than ER. Neither ER nor RR achieves comparable final performance to RTW, with both methods showing maximum success rates of only about 20\% even after completing the full training process.

\begin{figure}[tp]
    \centering
    \includegraphics[width=0.95\columnwidth]{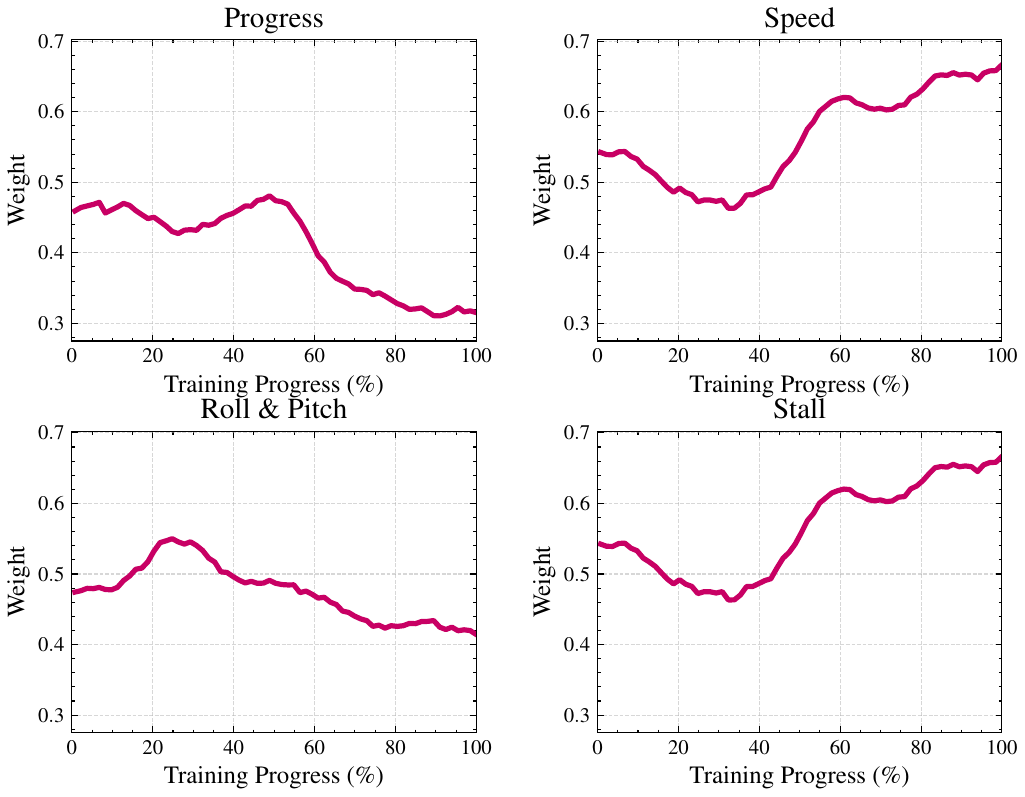}
    \caption{Evolution of auxiliary reward weights during training in Off-Road Vehicle Mobility.}
    \label{fig:weight_evolution}
\vspace{-10pt}
\end{figure}
\paragraph{Auxiliary Weights Evolution}
Figure~\ref{fig:weight_evolution} illustrates how RTW dynamically adapts auxiliary reward weights throughout training in the Off-Road Mobility task, revealing a systematic progression that resembles a curriculum. The progress reward weight initially remains stable (0.45) during early learning but decreases substantially (to 0.32) in later stages, indicating that once basic navigation is mastered, the teacher reduces emphasis on simple forward movement. Concurrently, the roll-pitch penalty weight follows an inverted U-shaped pattern, first increasing (0.47 to 0.55) to emphasize stability during skill acquisition, then gradually decreasing (to 0.42) as the vehicle develops robust control strategies. This pattern exemplifies RTW's "training wheels" approach—first prioritizing fundamental stability, then gradually reducing this constraint as competence develops. Most notably, both speed and stall prevention weights show significant increases (0.52 to 0.65 for speed; 0.53 to 0.65 for stall prevention) in the latter half of training, reflecting a shift toward performance optimization once basic skills are mastered.

A particularly interesting pattern emerges in the relationship between speed and stall prevention weights, which follow remarkably similar trajectories despite being separate auxiliary components. This synchronized adaptation suggests that RTW has discovered the intrinsic relationship between these components—both fundamentally encourage the vehicle to maintain forward momentum. While a human designer might treat these as distinct concerns, RTW autonomously recognizes their functional similarity and adjusts them in tandem. This emergence of correlated weight adjustments demonstrates RTW's ability to identify underlying relationships between superficially different reward components without explicit programming, further validating the system's capacity to discover meaningful reward structures adaptively.

\subsection{Physical Demonstration}
\label{subsec:phy_results}
\begin{figure}[ht]
    \centering
    \includegraphics[width=\columnwidth]{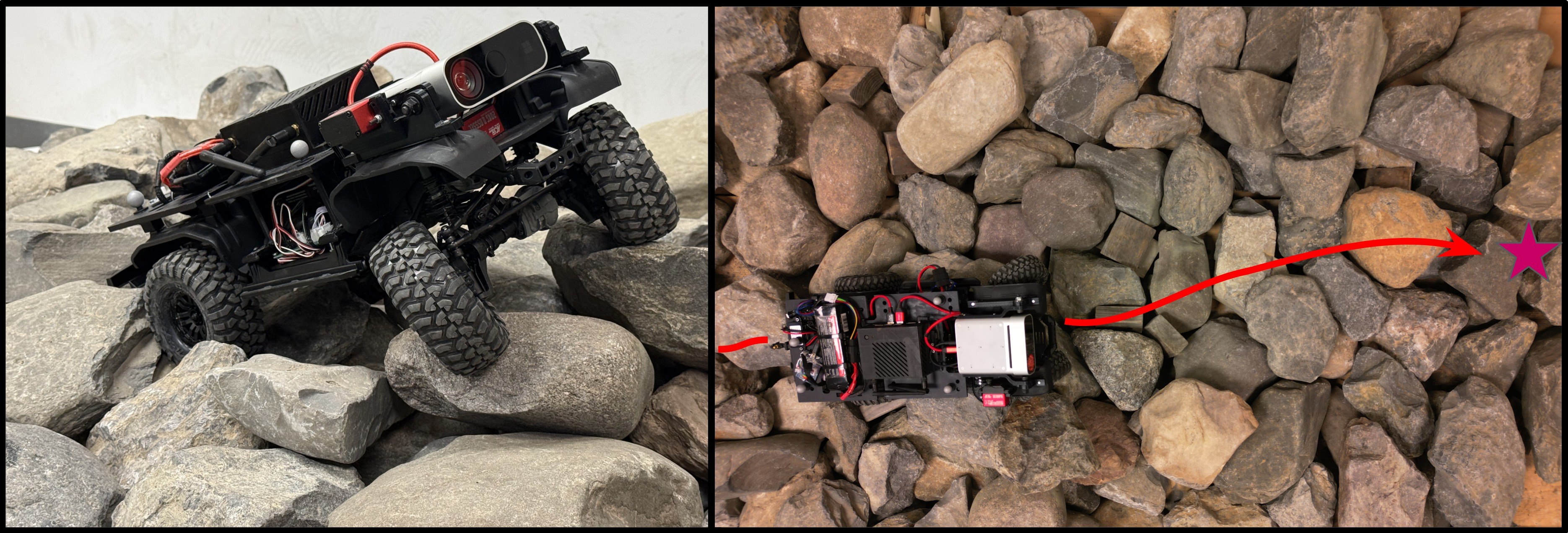}
    \caption{Physical Off-Road Testbed for RTW.}
    \label{fig::real}
\end{figure}

To validate the effectiveness of our framework, we deploy the RL policies, i.e., RTW, ER and RR learned in simulation, on a physical 1/10th scale open-source Verti-4-Wheeler robot~\cite{datar2024toward} platform on an off-road mobility testbed constructed by rocks (Fig.~\ref{fig::real}). The robot is a four-wheeled platform based on an off-the-self, two-axle, four-wheel-drive, off-road vehicle from Traxxas. The onboard computation platform is a NVIDIA Jetson Xavier NX module. The results are presented in  Table~\ref{tab::results}. 

\begin{table}
\caption{Experiment Results of \textsc{RTW}, \textsc{ER} and \textsc{RR}: success rate, mean traversal time (of successful trials), and mean absolute roll and pitch angles.}
\centering
\resizebox{\columnwidth}{!}{%
\small
\setlength{\tabcolsep}{4pt}
\begin{tabular}{ccccccccc}
\toprule
                    & {\textsc{RTW}} & {\textsc{ER}} & {\textsc{RR}} \\
\midrule
{Success Rate $\uparrow$}  & {\textbf{5/5}} & {2/5} & {1/5}\\
{Traversal Time $\downarrow$} & \textbf{{6.05}}$\pm$0.95 & {6.48}$\pm$0.58 & {8.68}$\pm$2.88\\
{Absolute Roll $\downarrow$}      & \textbf{{4.8}\textdegree}$\pm$5.01\textdegree & {9.12\textdegree$\pm$7.35\textdegree} & {7.82\textdegree$\pm$9.61\textdegree}\\
{Absolute Pitch $\downarrow$}      & \textbf{{7.09\textdegree}}$\pm$3.71\textdegree & {7.33\textdegree$\pm$4.91\textdegree} & {8.99\textdegree$\pm$7.70\textdegree}\\
\bottomrule
\end{tabular}%
}
\label{tab::results}
\vspace{-10pt}
\end{table}

The physical validation results presented in Table~\ref{tab::results} demonstrate the clear superiority of the RTW approach compared to ER and RR across all performance metrics.
Most notably, RTW achieves a perfect success rate (5/5 trials), substantially outperforming both ER (2/5) and RR (1/5). This marked improvement in reliability can be attributed to RTW's dynamic auxiliary reward weighting mechanism described in Section~\ref{sec:approach}, which adaptively emphasizes different components of the task based on the agent's current learning stage and performance history.

The temporal efficiency of RTW is evidenced by the shortest mean traversal time compared to ER and RR. 
Also, the vehicle stability metrics further validate RTW's effectiveness in maintaining control during navigation tasks. 
RTW demonstrates improved vehicle stability through significantly reduced orientation angles. For absolute roll, RTW achieves 4.80° compared to ER's 9.12° (47.4\% reduction) and RR's 7.82° (38.6\% reduction). For absolute pitch, RTW's 7.09° outperforms ER's 7.33° (3.3\% reduction) and RR's 8.99° (21.1\% reduction). These improvements confirm RTW's enhanced control capabilities during navigation tasks.

These results align with the theoretical foundations of our approach outlined in Section~\ref{sec:approach}. The teacher agent in RTW effectively leverages its state representation—comprising historical weight configurations, primary rewards, and auxiliary reward vectors—to generate optimal weight distributions that guide the student agent through the learning process. The decomposed reward structure allows for targeted emphasis on specific components of the task, adapting as the agent's capabilities evolve.

%% file: contents/conclusions.tex
\section{Conclusions and Future Work}
\label{sec::conclusions}

In this paper, we address the persistent challenge of designing effective auxiliary rewards for RL in complex robotics tasks.  The reliance on manual reward engineering is time-consuming, prone to bias, and often fails to adapt to the robot's changing capabilities during training. To overcome these limitations, we introduce RTW, a novel teacher-student framework that automatically adapts auxiliary reward weights throughout the learning process. Through extensive simulation and physical experiments in challenging confined-space navigation and off-road vehicle mobility tasks, we demonstrate that RTW outperforms expert-designed rewards, reward randomization, and the classical MPPI planner. Notably, RTW demonstrated an interesting capability to identify functional relationships between auxiliary rewards, automatically synchronizing conceptually related components like speed and stall prevention without explicit programming.

Future work can focus on extending RTW to automatically discover and incorporate new auxiliary reward components, potentially leveraging techniques from unsupervised learning or intrinsic motivation.  Investigating more sophisticated teacher agent architectures, such as those incorporating attention mechanisms or recurrent networks, can further improve the teacher's ability to model long-term dependencies in the student's learning process.  Exploring the application of RTW to a broader range of robotic tasks, including manipulation and human-robot interaction, is also a promising direction.